\documentclass[conference]{IEEEtran}
\IEEEoverridecommandlockouts
\usepackage{cite}
\usepackage{amsmath,amssymb,amsfonts}
\usepackage{algorithmic}
\usepackage{graphicx}
\usepackage{textcomp}
\usepackage{xcolor}
\usepackage{booktabs}
\usepackage[ruled,linesnumbered]{algorithm2e}
\def\BibTeX{{\rm B\kern-.05em{\sc i\kern-.025em b}\kern-.08em
    T\kern-.1667em\lower.7ex\hbox{E}\kern-.125emX}}
\begin{document}

\title{AOSR-Net: All-in-One Sandstorm Removal Network\\
}

% \and
% \and
% \and
% \and
% \and
\author{\IEEEauthorblockN{Yazhong Si$^{1,2\dagger}$, Xulong Zhang$^{1\dagger}$\thanks{$^\dagger$ Both authors have made equal contributions.}, Fan Yang$^{2}$, Jianzong Wang$^{1\ast}$\thanks{$^{\ast}$Corresponding author: Jianzong Wang (jzwang@188.com)},  Ning Cheng$^{1}$, Jing Xiao$^{1}$}
\IEEEauthorblockA{\textit{$^{1}$Ping An Technology (Shenzhen) Co., Ltd.} \\ 
\textit{$^{2}$Hebei University of Technology}
}
}

\maketitle

\begin{abstract}
Most existing sandstorm image enhancement methods are based on traditional theory and prior knowledge, which often restrict their applicability in real-world scenarios. In addition, these approaches often adopt a strategy of color correction followed by dust removal, which makes the algorithm structure too complex. To solve the issue, we introduce a novel image restoration model, named \emph{all-in-one sandstorm removal network (AOSR-Net)}. This model is developed based on a re-formulated sandstorm scattering model, which directly establishes the image mapping relationship by integrating intermediate parameters. Such integration scheme effectively addresses the problems of over-enhancement and weak generalization in the field of sand dust image enhancement. Experimental results on synthetic and real-world sandstorm images demonstrate the superiority of the proposed AOSR-Net over state-of-the-art (SOTA) algorithms.
\end{abstract}

\begin{IEEEkeywords}
Sandstorm image, Sandstorm scattering model, Convolutional neural networks, Image restoration
\end{IEEEkeywords}

\section{Introduction}
Images captured in sandstorm weather are often affected by Mie scattering, resulting in low visibility, color degradation, and halo artifacts \cite{Zhi-Vis}. These issues not only diminish visual perception but significantly impact high-level image processing tasks. Sandstorm enhancement techniques are especially important for minimizing the loss of system performance caused by degraded images. However, sandstorm removal is still a quite challenging ill-posed issue due to the inhomogeneous scattering of visible light by sand particles, and dust particles will introduce significant noise resulting in degraded images lacking detailed textures.

Current research on sandstorm image processing can be broadly categorized into traditional-based methods, model-based methods, and learning-based methods.

The commonly utilized traditional techniques include gamma correction \cite{kenk2020visibility,jeon2022sand}, Retinex theory \cite{alluhaidan2019retinex,GAO2021165659,zhang2022retinex} and CLAHE \cite{shi2019let,hua2022colour,shi2020normalised}. While traditional enhancement algorithms can enhance the visual perception of images to some degree, the generalization ability of such algorithms is too weak and can only be applied to certain specific sanddust scenes. Furthermore, due to over-enhancement, these algorithms often cause some issues such as artifacts and secondary color distortion.

The atmospheric scattering model \cite{model} is commonly employed in image dehazing research and can be written as:

\begin{equation}
	I(x) = A[1 - t(x)] + J(x)t(x)
	\label{eq:1}
\end{equation}
here $I(x)$ denotes the foggy image; $J(x)$ indicates the clear reference; $t(x)$ and $A$ are the intermediate parameters of the model. Currently, Dark channel prior (DCP) theory \cite{he2010single} and the variants \cite{lee2022eximious,peng2018generalization,shi2023sand,lee2022efficient} are the widely accepted way for estimating the intermediate parameters.

\begin{figure}
	\centering
	\includegraphics[width=0.49\textwidth]{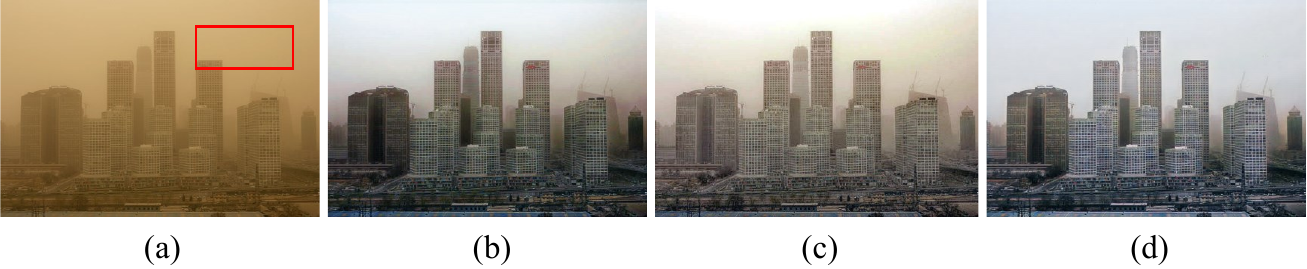}
	\caption{Visual comparison on real sandstorm image. (a) Sandstorm image; (b) FS \cite{si2022sand}; (c) ROP$^+$ \cite{liu2022rank}; (d) AOSR-Net.}
	\label{fig.1}
\end{figure}

\begin{figure*}
	\centering
	\includegraphics[width=0.98\textwidth]{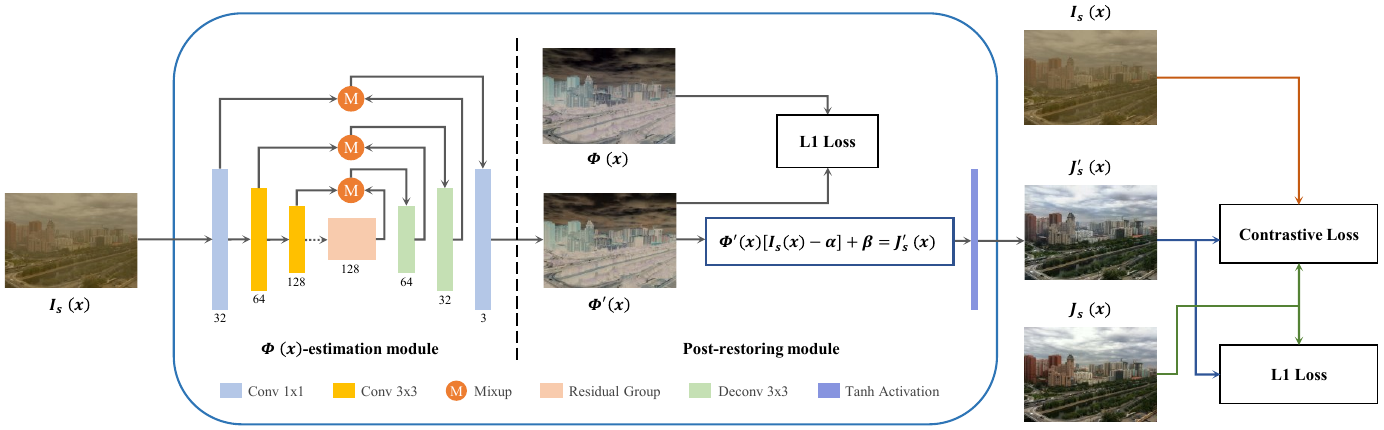}
	\caption{The framework of AOSR-Net, which consists of two key components: the $\Phi(x)$-estimation module and the post-restoring module. Please refer to the text for a more in-depth explanation.}
	\label{fig.2}	
\end{figure*}

The degradation characteristics of sandstorm images make most model-based sandstorm restoration methods \cite{shi2022sand,liu2021rank,dhara2020color,lee2021efficient,lee2022sandstorm} utilized the step-by-step strategy i.e., they first pre-process the degraded image with color correction and then use the physical model to eliminate the dust by following the process of image dehazing. Based on DCP theory, Shi et al. \cite{shi2022sand} introduced a sandstorm removal method, which is composed of red channel-based correction (RCC) and blue channel-based dust particle removal (BDPR) two parts. In their method, RCC is first used to balance the degraded color in the red channel and then fused the corrected image with the feature map processed by BDPR. Dhara et al. \cite{dhara2020color} proposed a cast-adaptive nonlinear transformation for color correcting and introduced a cast-adaptive airlight refinement method to ensure the results with no halo artifacts. The DCP theory will fail for sandstorm images, to solve the issue, Gao et al. \cite{gao2020sand} designed a reversed blue channel prior (RBCP) to calculate middle parameters and restore the clear image.

However, there are still some issues in existing sand dust image restoration algorithms that cannot be ignored. Specifically, the prior knowledge based on mathematical statistics is inherently surrounded by some fixed uncertainties. In addition, the physical model is designed for the haze image imaging process, and it is far-fetched to directly apply the model in sand dust image processing tasks.

Recently, the remarkable success of convolutional neural networks (CNN) \cite{si2022novel,gao2023very,Tang2023QI,Zhang2023Dynamic,Wu2022Pose,deng2023PMVC} has led to an increasing interest of the researchers in learning-based sandstorm image reconstruction algorithms \cite{huang2022sidnet,si2022sand,huang2020image,huang2021ffnet}. Inspired by the previous research, Si et al. \cite{si2022sand} introduced a fusion strategy (FS) to improve the visibility of sandstorm images, which combined the traditional color correction algorithm with learning-based CNN to ensure the stability and generalizability of the approach. Huang et al. \cite{huang2020image} presented a synthetic strategy for getting paired training sandstorm images. Following the benchmark, they proposed an image dedusting network with feature fusion \cite{huang2021ffnet}. However, the lack of adequate consideration of the imaging process in the benchmark makes the algorithm inapplicable to real-world sandstorm scenarios.

Although the previous works have achieved certain advancements, the intricate distribution characteristics of sandstorm images posed significant challenges still worth further exploration, such as over-enhancement and incomplete dust removal. To bridge the gap, we introduce a novel sandstorm image restoration method in this paper. Fig.\ref{fig.1} provides a visual comparison in real-world sandstorm image.

The main contributions of the paper can be outlined as:
\begin{itemize}
\item We integrate the intermediate parameters of the sandstorm imaging model into an unknown variable and construct a back-propagation model for restoring the clear images.
\item We propose a simple yet efficient CNN model, named all-in-one sandstorm removal network (AOSR-Net), which is designed to estimate the integrated variable in the back-propagation model.
\end{itemize}

\section{Proposed Method}
% In this section, the details of the proposed method are explained. We first introduce the sand dust scattering model and then reverse the imaging process to construct a back-propagation model. Finally, we will discuss the structure of the proposed AOSR-Net.

\subsection{Transformed formula}
To promote the research of model-based sand dust image restoration algorithm, Si et al. \cite{si2022comprehensive} conducted an extensive perceptual investigation on real-world sandstorm images and introduced a sand-dust scattering model, represented as:
\begin{equation}
    {I_s}(x) = {A_s} + \left[ {{J_s}(x) - {{A'_s}}} \right]t_s(x)
    \label{eq:2}
\end{equation}
\begin{equation}
    t_s(x)=e^{-\beta d_s(x)}
    \label{eq:3}
\end{equation}
here $I_s(x)$ pertains sandstorm image, $J_s(x)$ signifies clear image, $A_s$ denotes global color deviation value, and $A'_s$ corresponds to complementary color of $A_s$. Additionally, $t_s(x)$ represents transmission map, $\beta$ stands for attenuation coefficient, and $d_s(x)$ pertains to scene depth.

Based on Eq.\ref{eq:2}, the clear image is restored by:
\begin{equation}
    J_s(x)=\frac{1}{t_s(x)}\left[I_s(x)-A_s\right]+A_s'
    \label{eq:4}
\end{equation}

As explained in the work \cite{li2017aod}, compared to estimating $t_s(x)$ and $A_s$ separately, directly calculating the integrated formula can yield smaller restoration errors at the pixel level. To this end, we re-formulate Eq.\ref{eq:4} as the following:
\begin{equation}
    J_s(x)=\Phi(x)\big[I_s(x)-\alpha\big]+\beta
    \label{eq:5}
\end{equation}
\begin{equation}
    \Phi(x)=\dfrac{I_s(x)-A_s+t_s(x)\big(A_s^\prime-\beta\big)}{t_s(x)\big[I_s(x)-\alpha\big]}
    \label{eq:6}
\end{equation}
where $\Phi(x)$ is the integrated variable; $\alpha$ and $\beta$ are the control coefficients employed to normalize the pixel interval of $\Phi(x)$.

In that way, we simplified the back-propagation model by integrating the potential computational relationships into $\Phi(x)$. From Eq.\ref{eq:6}, we find that there is some kind of transformation relationship between $\Phi(x)$ and $I_s(x)$, so we aim to build such a mapping process in a data-driven way and then restore the clear image according to Eq.\ref{eq:5}.

\subsection{Network architecture}
Fig.\ref{fig.2} illustrates the framework of AOSR-Net, which is comprised of two main components: $\Phi(x)$-estimation module, and the post-restoring module. As displayed in Fig.\ref{fig.2}, the $\Phi(x)$-estimation module adopts a simple encoder–decoder structure, it comprises four convolutional layers and two deconvolutional layers, with a default PReLu activation layer following each convolutional/deconvolutional calculation. To boost the nonlinear representation ability of the network, we embed a residual group composed of three residual blocks \cite{he2016} cascaded in high-dimensional latent space. 

In addition, we implement the mixup technique\cite{zhang2017mixup} in AOSR-Net to enhance its ability of adaptively fuse the shallow and deep information, thereby improving the network's generalization capabilities. We describe the mixup operation as follows:
\begin{equation}
    f = mix\left( {{f_1},{f_2}} \right) = \sigma \left( \xi  \right) * {f_1} + \left[ {1 - \sigma \left( \xi  \right)} \right] * {f_2}
    \label{eq:7}
\end{equation}
where $f$ is the fusion data; $f_1$, $f_2$ are the input features; $\sigma$ denotes the sigmoid activation; $\xi$ is the trainable factor, it can dynamically adjust the fusion proportion of $f_1$ and $f_2$. The mixup regularization has been verified that optimizing $f$ on mixed-up data results in better transformation capability of the network and improves model calibration \cite{thulasidasan2019mixup}.

% Actually, AOSR-Net is mainly designed for estimating $\Phi(x)$ in the proposed algorithm. At the post-processing part of the network, the clear image can be restored using only a linear equation formula and a nonlinear activation function.

\subsection{Loss function}
In this paper, we employ L1-loss and contrastive learning loss \cite{wu2021contrastive} to optimize the proposed AOSR-Net at the pixel and perceptual levels, respectively. Below we give the details form of the sub-loss function.
\subsubsection{L1-loss}
As shown in Fig.\ref{fig.2}, we calculate the pixel errors on the two output nodes of the network to guide it to estimate $\Phi(x)$ accurately, which can be described in the form:
\begin{equation}
    L_1=\sum\limits_{i=1}^N\left\|\Phi_i(x)-\Phi'_i(x)\right\|_1+\left\|J_{si}(x)-J'_{si}(x)\right\|_1
    \label{eq:8}
\end{equation}
here $\Phi'(x)$ and $J^{'}_s(x)$ represent the estimated outcomes; while $\Phi(x)$ and $J_s(x)$ correspond to the objective ground truth (GT); $N$ signifies the number of training samples.

\subsubsection{Contrastive learning loss}
Recently, contrastive learning is gradually sparking the interest of researchers, \cite{wu2021contrastive,Tang2023VQ}, it aims to acquire a representation that closely aligns with the 'positive' distribution while distancing itself from the 'negative' distribution. We adopt the contrastive learning loss \cite{wu2021contrastive} to minimize the perceptual error of the network. Mathematically, the perceptual sub-loss can be formulated as:
\begin{equation}
    L_c=\sum_{j=1}^{n}\omega_{j}\cdot\dfrac{D\Big[G_{j}\big(J(x)\big),G_{j}\big({J}^{'}(x)\big)\Big]}{D\Big[G_{j}\big(I(x)\big),G_{j}\big({J}^{'}(x)\big)\Big]}
    \label{eq:9}
\end{equation}
where $D(x,y)$ is the $L_1$ distance between $x$ and $y$ in representation space; $G_i$, $i$ = 1, 2, $\cdots n_d$ represents the latent features extracted from the $i$-th layer of the pre-trained model \cite{simonyan2014very}, which is held constant during the process. The total objective function of AOSR-Net is:
\begin{equation}
    L_{total}=\lambda_{1}L_1+\lambda_{2}L_c
    \label{eq:10}
\end{equation}
where $\lambda_2$ are hyperparameters, and their specific values are configured as 0.25 and 0.5, respectively.

\section{Experiments}\label{sec3}
\subsection{Training dataset}
\begin{figure}
	\centering
	\includegraphics[width=0.49\textwidth]{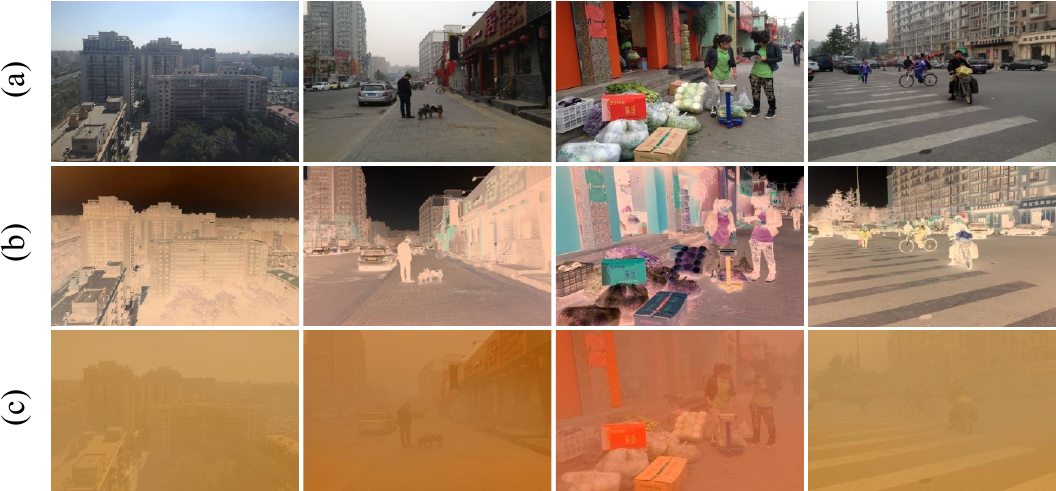}
	\caption{Example of the synthesized images. (a) Original images; (b) Objective integrated variables; (c) Synthesized sandstorm images.}
	\label{fig.3}
\end{figure}

Dataset generally play a vital role for the learning-based sandstorm image restoration methods. However, it is practically impossible to simultaneously obtain both real-world sandstorm images and the corresponding GT. To bridge the gap, Si et al. \cite{si2022comprehensive} introduced a synthesis strategy through a comprehensive analysis of the imaging process of sand dust images. 
\begin{algorithm}
    \caption{Steps for synthesizing the training data}
    \label{alg:1}
    \KwIn{$J_s(x)$}
    \KwOut{$I_s(x)$, $\Phi(x)$}
        \SetKwFor{For}{Calculate $t_s(x)$:}{}{}
        \For{}{
            Select $\beta$ from [0.3, 0.6];\\
            Estimate $d_s(x)$ based on \cite{estdeep};\\
            Compute $t_s(x)$ using Eq.\ref{eq:3};\\
        }
        \SetKwFor{For}{Get the intermediate values:}{}{}
        \For{}{
            Select $A_s$;\\
            Calculate $A^{'}_s$: $A^{'}_s$=1-$A_s$;\\
        }
        \SetKwFor{For}{Synthesize the sandstorm images:}{}{}
        \For{}{
            Synthesize $I_s(x)$ according to Eq.\ref{eq:2}.
        }
        \SetKwFor{For}{Calculate the objective integrated variable $\Phi(x)$:}{}{}
        \For{}{
            Calculate $\Phi(x)$ according to Eq.\ref{eq:6}.
        }
\end{algorithm}
Following the method \cite{si2022comprehensive}, we synthesized a training dataset containing 4000 sets of images, each set including a sandstorm image and its corresponding objective integrated variable and GT. The steps of data synthesis are described in Algorithm \ref{alg:1}. We start by randomly sampling the attenuation coefficient $\beta$ from a uniform distribution within the range of [0.3, 0.6] and adopt \cite{estdeep} to estimate the scene depth $d_s(x)$. Then we randomly select $A_s$ from the color library constructed by \cite{si2022comprehensive} and calculate the corresponding complementary color $A^{'}_s$. Finally, the sandstorm image $I_s(x)$ and the corresponding objective fusion variables $\Phi(x)$ are synthesized by Eq.\ref{eq:2} and Eq.\ref{eq:6}, respectively. Example of the synthesized images as illustrated in Fig.\ref{fig.3}.
\begin{figure*}
	\centering
	\includegraphics[width=0.98\textwidth]{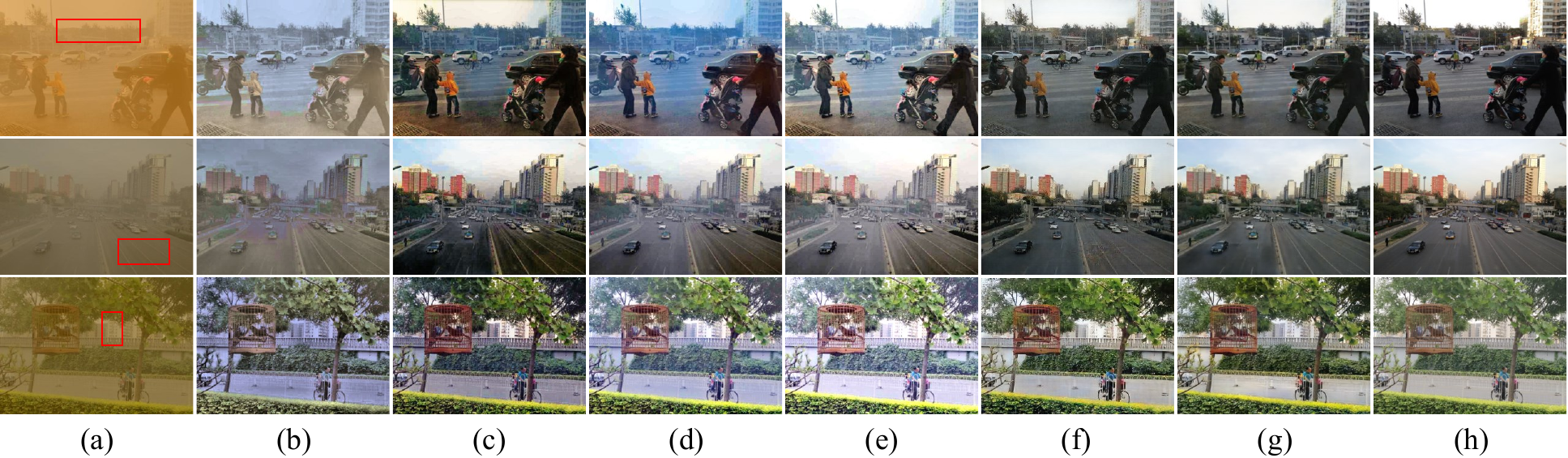}
	\caption{Visual comparison on synthetic sandstorm images. (a) Sandstorm image; (b) CBCS \cite{hua2022colour}; (c) FS \cite{si2022sand}; (d) ROP \cite{liu2021rank}; (e) ROP$^+$ \cite{liu2022rank}; (f) Pix2pix-H \cite{si2022comprehensive}; (g) AOSR-Net; (h) GT.}
	\label{fig.4}
\end{figure*}
\subsection{Implementation details}
In the experiments, AOSR-Net is implemented using PyTorch 1.9.0 on an NVIDIA GeForce RTX2070 GPU. The model is iterated for 100 epochs utilizing the Adam optimizer, with an initial learning rate of $1\times10^{-3}$. In terms of image synthesis, all the original images are selected from RESIDE \cite{RESIDE} and cropped into $256\times256\times3$. To normal the pixel interval of $\Phi(x)$, the control coefficients $\alpha$ and $\beta$ are configured as 1.6 and 1, respectively. 

Next, we will compare our AOSR-Net with the SOTA algorithms including CBCS \cite{hua2022colour}, FS \cite{si2022sand}, ROP \cite{liu2021rank}, ROP$^+$ \cite{liu2022rank} and Pix2pix-H \cite{si2022comprehensive}. To comprehensively evaluate the performance of the algorithms, we utilize PSNR, SSIM, and CIEDE2000 as the full-reference metrics. Additionally, we employ NIQE \cite{niqe} and SSEQ \cite{sseq} as the non-reference metrics. The higher PSNR, and SSIM values and lower CIEDE2000, NIQE, and SSEQ values represent better performance.

\subsection{Comparison with SOTA methods}

\subsubsection{Results on synthetic sandstorm images}
To qualitatively and quantitatively evaluate the performance of the algorithms, we synthesized 100 pairs of images as the testing dataset. The comparison results of the algorithms as shown in Fig.\ref{fig.4}. One can see that CBCS \cite{hua2022colour} and ROP \cite{liu2021rank} fail to balance the degraded color, and their results are still visually abnormal. FS \cite{si2022sand} and ROP$^+$ \cite{liu2022rank} can correct the color deviation, but the over-enhancement results in obvious haloes and artifacts in their results. As an end-to-end algorithm, due to the lack of guiding training information, Pix2pix-H \cite{si2022comprehensive} produces numerous pseudo-textures in local areas, resulting in a serious loss of details. Compared with the existing algorithms, AOSR-Net can generate more natural results with richer detail information, which are more visually faithful to GT. Table \ref{tab:1} provides the comparison of the evaluation metrics, where the optimal and sub-optimal performance are bolded and underlined, respectively. In Table \ref{tab:1}, one can see that AOSR-Net achieves the best performance both in full-reference and non-reference evaluations.

\begin{table}
  \centering
  \caption{Evaluation metrics comparison of the algorithms}
  \setlength{\tabcolsep}{0.2mm}{
    \begin{tabular}{ccccccc}
    \toprule
    Metrics & CBCS\cite{hua2022colour}  & FS\cite{si2022sand}    & ROP\cite{liu2021rank}  & ROP$^+$\cite{liu2022rank}  & Pix2pix-H\cite{si2022comprehensive} & AOSR-Net \\
    \midrule
    PSNR  & 16.2341 & 20.294 & 19.1952 & 18.8778 & \underline{24.2853} & \textbf{24.6728} \\
    SSIM  & 0.5704 & 0.7464 & 0.6344 & 0.7018 & \underline{0.8142} & \textbf{0.8275} \\
    CIEDE2000 & 32.6602 & 30.3896 & 28.1177 & 24.4766 & \underline{18.6402} & \textbf{17.9832} \\
    NIQE  & 3.1846 & 3.6428 & \underline{3.1504} & 3.1683 & 3.2603 & \textbf{2.9625} \\
    SSEQ  & 25.1938 & 22.4138 & 24.0943 & 24.0783 & \underline{22.3911} & \textbf{21.5081} \\
    \bottomrule
    \end{tabular}}%
  \label{tab:1}%
\end{table}%

\begin{figure}
	\centering
	\includegraphics[width=0.49\textwidth]{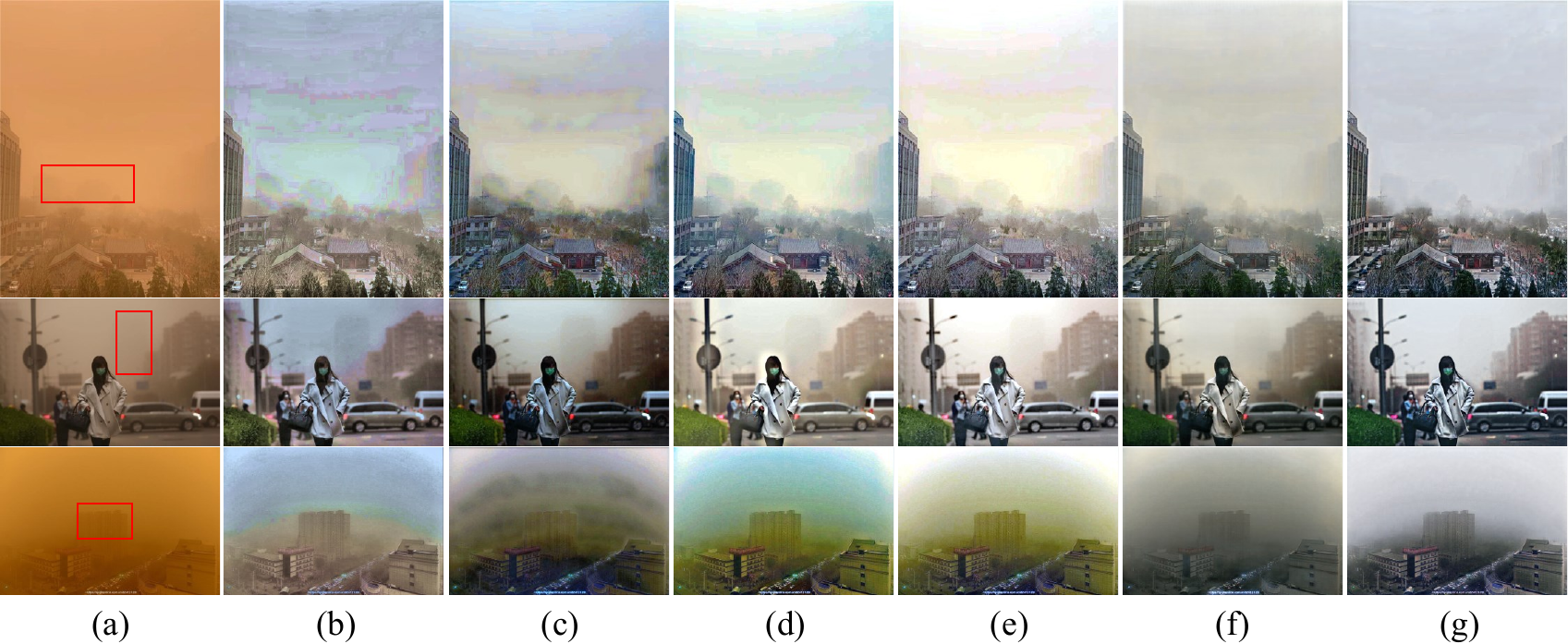}
	\caption{Visual comparisons on RSTS. (a) Sandstorm image; (b) CBCS \cite{hua2022colour}; (c) FS \cite{si2022sand}; (d) ROP \cite{liu2021rank}; (e) ROP$^+$ \cite{liu2022rank}; (f) Pix2pix-H \cite{si2022comprehensive}; (g) AOSR-Net.}
	\label{fig.5}
\end{figure}
\subsubsection{Results on real scenario sandstorm images}
Despite being trained on the synthetic dataset, AOSR-Net performs excellent generalization in real-scene sandstorm images. As shown in Fig.\ref{fig.5}, we evaluate the performance of the SOTA on Real-world sandstorm testing set (RSTS) \cite{si2022comprehensive}. In the top row of Fig.\ref{fig.5}, there is a significant blocking artifact in the result of CBCS \cite{hua2022colour}. While the color distortion issues still exist in the results of FS \cite{si2022sand}, ROP \cite{liu2021rank} and ROP$^+$ \cite{liu2022rank}. Pix2pix-H \cite{si2022comprehensive} can balance the color shifting, but the low contrast ratio causes its results to be visually dim and the local details are not prominent. As we can see from the last row of Fig.\ref{fig.5}, all comparison algorithms suffer from weak generalization, especially for such challenging sandstorm scenes, and they often fail with significant halos, artifacts, and color distortion in the results. Overall, AOSR-Net outperforms other sandstorm enhancement algorithms by effectively correcting the shifting colors and eliminating dust more thoroughly.

\section{Conclusion}
In this paper, we propose an all-in-one pipeline for sandstorm image restoration. Specifically, we integrate the intermediate parameters of the sand dust imaging model into an unknown variable and then design a CNN named AOSR-Net to estimate the integrated variable. Finally, we restore the clear images via the re-formulate sandstorm scattering model. Experimental results indicate that the performance of the proposed algorithm outperforms existing SOTA methods on both synthetic and real scenario sand dust images.

\section{Acknowledgement}
This paper is supported by the Key Research and Development Program of Guangdong Province under grant No.2021B0101400003. Corresponding author is Jianzong Wang from Ping An Technology (Shenzhen) Co., Ltd (jzwang@188.com).

\bibliographystyle{IEEEtran.bst}
\bibliography{mybib.bib}

\begin{thebibliography}{10}
\providecommand{\url}[1]{#1}
\csname url@samestyle\endcsname
\providecommand{\newblock}{\relax}
\providecommand{\bibinfo}[2]{#2}
\providecommand{\BIBentrySTDinterwordspacing}{\spaceskip=0pt\relax}
\providecommand{\BIBentryALTinterwordstretchfactor}{4}
\providecommand{\BIBentryALTinterwordspacing}{\spaceskip=\fontdimen2\font plus
\BIBentryALTinterwordstretchfactor\fontdimen3\font minus \fontdimen4\font\relax}
\providecommand{\BIBforeignlanguage}[2]{{%
\expandafter\ifx\csname l@#1\endcsname\relax
\typeout{** WARNING: IEEEtran.bst: No hyphenation pattern has been}%
\typeout{** loaded for the language `#1'. Using the pattern for}%
\typeout{** the default language instead.}%
\else
\language=\csname l@#1\endcsname
\fi
#2}}
\providecommand{\BIBdecl}{\relax}
\BIBdecl

\bibitem{Zhi-Vis}
N.~Zhi, S.~Mao, and M.~Li, ``Visibility restoration algorithm of dust-degraded images,'' \emph{Journal of Image and Graphics}, vol.~21, no.~12, pp. 1585--1592, 2016.

\bibitem{kenk2020visibility}
M.~A. Kenk, M.~Hassaballah, M.~A. Hameed, and S.~Bekhet, ``Visibility enhancer: adaptable for distorted traffic scenes by dusty weather,'' in \emph{2020 2nd Novel Intelligent and Leading Emerging Sciences Conference (NILES)}.\hskip 1em plus 0.5em minus 0.4em\relax IEEE, 2020, pp. 213--218.

\bibitem{jeon2022sand}
J.-J. Jeon, T.-H. Park, and I.-K. Eom, ``Sand-dust image enhancement using chromatic variance consistency and gamma correction-based dehazing,'' \emph{Sensors}, vol.~22, no.~23, p. 9048, 2022.

\bibitem{alluhaidan2019retinex}
M.~S. Alluhaidan, M.~Alsafasfeh, I.~Abdel-Qader, and O.~Abudayyeh, ``Retinex-based framework for visibility enhancement during inclement weather with tracking and estimating distance of vehicles,'' in \emph{2019 IEEE Jordan International Joint Conference on Electrical Engineering and Information Technology (JEEIT)}.\hskip 1em plus 0.5em minus 0.4em\relax IEEE, 2019, pp. 250--255.

\bibitem{GAO2021165659}
G.~Gao, H.~Lai, Y.~Liu, L.~Wang, and Z.~Jia, ``Sandstorm image enhancement based on yuv space,'' \emph{Optik}, vol. 226, p. 165659, 2021.

\bibitem{zhang2022retinex}
W.~Zhang, L.~Dong, and W.~Xu, ``Retinex-inspired color correction and detail preserved fusion for underwater image enhancement,'' \emph{Computers and Electronics in Agriculture}, vol. 192, p. 106585, 2022.

\bibitem{shi2019let}
Z.~Shi, Y.~Feng, M.~Zhao, E.~Zhang, and L.~He, ``Let you see in sand dust weather: A method based on halo-reduced dark channel prior dehazing for sand-dust image enhancement,'' \emph{IEEE Access}, vol.~7, pp. 116\,722--116\,733, 2019.

\bibitem{hua2022colour}
Z.~Hua, L.~Qi, M.~Guan, H.~Su, and Y.~Sun, ``Colour balance and contrast stretching for sand-dust image enhancement,'' \emph{IET Image Processing}, vol.~16, no.~14, pp. 3768--3780, 2022.

\bibitem{shi2020normalised}
Z.~Shi, Y.~Feng, M.~Zhao, E.~Zhang, and L.~He, ``Normalised gamma transformation-based contrast-limited adaptive histogram equalisation with colour correction for sand--dust image enhancement,'' \emph{IET Image Processing}, vol.~14, no.~4, pp. 747--756, 2020.

\bibitem{model}
E.~J. McCartney, ``Optics of the atmosphere: scattering by molecules and particles,'' \emph{New York}, 1976.

\bibitem{he2010single}
K.~He, J.~Sun, and X.~Tang, ``Single image haze removal using dark channel prior,'' \emph{IEEE transactions on pattern analysis and machine intelligence}, vol.~33, no.~12, pp. 2341--2353, 2010.

\bibitem{lee2022eximious}
H.~S. Lee, ``Eximious sandstorm image improvement using image adaptive ratio and brightness-adaptive dark channel prior,'' \emph{Symmetry}, vol.~14, no.~7, p. 1334, 2022.

\bibitem{peng2018generalization}
Y.-T. Peng, K.~Cao, and P.~C. Cosman, ``Generalization of the dark channel prior for single image restoration,'' \emph{IEEE Transactions on Image Processing}, vol.~27, no.~6, pp. 2856--2868, 2018.

\bibitem{shi2023sand}
F.~Shi, Z.~Jia, H.~Lai, N.~K. Kasabov, S.~Song, and J.~Wang, ``Sand-dust image enhancement based on light attenuation and transmission compensation,'' \emph{Multimedia Tools and Applications}, vol.~82, no.~5, pp. 7055--7077, 2023.

\bibitem{lee2022efficient}
H.-S. Lee, ``Efficient color correction using normalized singular value for duststorm image enhancement,'' \emph{J}, vol.~5, no.~1, pp. 15--34, 2022.

\bibitem{si2022sand}
Y.~Si, F.~Yang, and Z.~Liu, ``Sand dust image visibility enhancement algorithm via fusion strategy,'' \emph{Scientific Reports}, vol.~12, no.~1, p. 13226, 2022.

\bibitem{liu2022rank}
J.~Liu, R.~W. Liu, J.~Sun, and T.~Zeng, ``Rank-one prior: Real-time scene recovery,'' \emph{IEEE Transactions on Pattern Analysis and Machine Intelligence}, 2022.

\bibitem{shi2022sand}
F.~Shi, Z.~Jia, H.~Lai, S.~Song, and J.~Wang, ``Sand dust images enhancement based on red and blue channels,'' \emph{Sensors}, vol.~22, no.~5, p. 1918, 2022.

\bibitem{liu2021rank}
J.~Liu, W.~Liu, J.~Sun, and T.~Zeng, ``Rank-one prior: Toward real-time scene recovery,'' in \emph{Proceedings of the IEEE/CVF Conference on Computer Vision and Pattern Recognition}, 2021, pp. 14\,802--14\,810.

\bibitem{dhara2020color}
S.~K. Dhara, M.~Roy, D.~Sen, and P.~K. Biswas, ``Color cast dependent image dehazing via adaptive airlight refinement and non-linear color balancing,'' \emph{IEEE Transactions on Circuits and Systems for Video Technology}, vol.~31, no.~5, pp. 2076--2081, 2020.

\bibitem{lee2021efficient}
H.~S. Lee, ``Efficient sandstorm image enhancement using the normalized eigenvalue and adaptive dark channel prior,'' \emph{Technologies}, vol.~9, no.~4, p. 101, 2021.

\bibitem{lee2022sandstorm}
H.~Lee, ``Sandstorm image enhancement using image-adaptive eigenvalue and brightness-adaptive dark channel network,'' \emph{Symmetry}, vol.~14, no.~11, p. 2310, 2022.

\bibitem{gao2020sand}
G.~Gao, H.~Lai, Z.~Jia, Y.~Liu, and Y.~Wang, ``Sand-dust image restoration based on reversing the blue channel prior,'' \emph{IEEE Photonics Journal}, vol.~12, no.~2, pp. 1--16, 2020.

\bibitem{si2022novel}
Y.~Si, F.~Yang, and N.~Chong, ``A novel method for single nighttime image haze removal based on gray space,'' \emph{Multimedia Tools and Applications}, vol.~81, no.~30, pp. 43\,467--43\,484, 2022.

\bibitem{gao2023very}
D.~Gao and D.~Zhou, ``A very lightweight and efficient image super-resolution network,'' \emph{Expert Systems with Applications}, vol. 213, p. 118898, 2023.

\bibitem{Tang2023QI}
H.~Tang, X.~Zhang, J.~Wang, N.~Cheng, and J.~Xiao, ``Qi-tts: Questioning intonation control for emotional speech synthesis,'' in \emph{ICASSP 2023 - 2023 IEEE International Conference on Acoustics, Speech and Signal Processing (ICASSP)}, 2023, pp. 1--5.

\bibitem{Zhang2023Dynamic}
X.~Zhang, H.~Tang, J.~Wang, N.~Cheng, J.~Luo, and J.~Xiao, ``Dynamic alignment mask ctc: Improved mask ctc with aligned cross entropy,'' in \emph{ICASSP 2023 - 2023 IEEE International Conference on Acoustics, Speech and Signal Processing (ICASSP)}, 2023, pp. 1--5.

\bibitem{Wu2022Pose}
J.~Wu, S.~Si, J.~Wang, X.~Qu, and X.~Jing, ``Pose guided human image synthesis with partially decoupled gan,'' in \emph{Asian Conference on Machine Learning}, ser. Proceedings of Machine Learning Research, vol. 189.\hskip 1em plus 0.5em minus 0.4em\relax PMLR, 2022, pp. 1133--1148.

\bibitem{deng2023PMVC}
Y.~Deng, H.~Tang, X.~Zhang, J.~Wang, N.~Cheng, and J.~Xiao, ``Pmvc: Data augmentation-based prosody modeling for expressive voice conversion,'' in \emph{31st ACM International Conference on Multimedia}, 2023.

\bibitem{huang2022sidnet}
J.~Huang, H.~Xu, G.~Liu, C.~Wang, Z.~Hu, and Z.~Li, ``Sidnet: A single image dedusting network with color cast correction,'' \emph{Signal Processing}, vol. 199, p. 108612, 2022.

\bibitem{huang2020image}
J.~Huang, Z.~Li, and C.~Wang, ``Image dust storm synthetic method based on optical model,'' in \emph{Machine Learning for Cyber Security: Third International Conference, ML4CS 2020, Guangzhou, China, October 8--10, 2020, Proceedings, Part III 3}.\hskip 1em plus 0.5em minus 0.4em\relax Springer, 2020, pp. 215--226.

\bibitem{huang2021ffnet}
J.~Huang, Z.~Li, C.~Wang, Z.~Yu, and X.~Cao, ``Ffnet: A simple image dedusting network with feature fusion,'' \emph{Concurrency and Computation: Practice and Experience}, vol.~33, no.~24, p. e6462, 2021.

\bibitem{si2022comprehensive}
Y.~Si, F.~Yang, Y.~Guo, W.~Zhang, and Y.~Yang, ``A comprehensive benchmark analysis for sand dust image reconstruction,'' \emph{Journal of Visual Communication and Image Representation}, vol.~89, p. 103638, 2022.

\bibitem{li2017aod}
B.~Li, X.~Peng, Z.~Wang, J.~Xu, and D.~Feng, ``Aod-net: All-in-one dehazing network,'' in \emph{Proceedings of the IEEE international conference on computer vision}, 2017, pp. 4770--4778.

\bibitem{he2016}
K.~He, X.~Zhang, S.~Ren, and J.~Sun, ``Deep residual learning for image recognition,'' in \emph{Proceedings of the IEEE conference on computer vision and pattern recognition}, 2016, pp. 770--778.

\bibitem{zhang2017mixup}
H.~Zhang, M.~Cisse, Y.~N. Dauphin, and D.~Lopez-Paz, ``mixup: Beyond empirical risk minimization,'' \emph{arXiv preprint arXiv:1710.09412}, 2017.

\bibitem{thulasidasan2019mixup}
S.~Thulasidasan, G.~Chennupati, J.~A. Bilmes, T.~Bhattacharya, and S.~Michalak, ``On mixup training: Improved calibration and predictive uncertainty for deep neural networks,'' \emph{Advances in Neural Information Processing Systems}, vol.~32, 2019.

\bibitem{wu2021contrastive}
H.~Wu, Y.~Qu, S.~Lin, J.~Zhou, R.~Qiao, Z.~Zhang, Y.~Xie, and L.~Ma, ``Contrastive learning for compact single image dehazing,'' in \emph{Proceedings of the IEEE/CVF Conference on Computer Vision and Pattern Recognition}, 2021, pp. 10\,551--10\,560.

\bibitem{Tang2023VQ}
H.~Tang, X.~Zhang, J.~Wang, N.~Cheng, and J.~Xiao, ``Vq-cl: Learning disentangled speech representations with contrastive learning and vector quantization,'' in \emph{ICASSP 2023 - 2023 IEEE International Conference on Acoustics, Speech and Signal Processing (ICASSP)}, 2023, pp. 1--5.

\bibitem{simonyan2014very}
K.~Simonyan and A.~Zisserman, ``Very deep convolutional networks for large-scale image recognition,'' \emph{arXiv preprint arXiv:1409.1556}, 2014.

\bibitem{estdeep}
F.~Liu, C.~Shen, G.~Lin, and I.~Reid, ``Learning depth from single monocular images using deep convolutional neural fields,'' \emph{IEEE transactions on pattern analysis and machine intelligence}, vol.~38, no.~10, pp. 2024--2039, 2015.

\bibitem{RESIDE}
B.~Li, W.~Ren, D.~Fu, D.~Tao, D.~Feng, W.~Zeng, and Z.~Wang, ``Benchmarking single-image dehazing and beyond,'' \emph{IEEE Transactions on Image Processing}, vol.~28, no.~1, pp. 492--505, 2018.

\bibitem{niqe}
A.~Mittal, R.~Soundararajan, and A.~C. Bovik, ``Making a “completely blind” image quality analyzer,'' \emph{IEEE Signal processing letters}, vol.~20, no.~3, pp. 209--212, 2012.

\bibitem{sseq}
L.~Liu, B.~Liu, H.~Huang, and A.~C. Bovik, ``No-reference image quality assessment based on spatial and spectral entropies,'' \emph{Signal processing: Image communication}, vol.~29, no.~8, pp. 856--863, 2014.

\end{thebibliography}

\end{document}